\documentclass[letterpaper, 10 pt, conference]{ieeeconf}  % Comment this line
\IEEEoverridecommandlockouts                              % This command is only needed if
\small
\title{Negotiation-Aware Reachability-Based Safety Verification for Autonomous Driving in Interactive Scenarios}
\normalsize

\author{Ran Tian, Anjian Li, Masayoshi Tomizuka, and Liting Sun
    \thanks{
         This work is presented at the ICRA 2021 Workshop on Safe Robot Control with Learned Motion and Environment Models.
     }
     \thanks{
         Ran Tian, Masayoshi Tomizuka, and Liting Sun are with University of California, Berkeley. ({\{rantian, tomizuka, litingsun\}@berkeley.edu}).
     }
     \thanks{
         Anjian Li is with Simon Fraser University, Burnaby, BC, Canada, V5A 1S6 ({anjianl@sfu.ca}).
     }
}
\usepackage{tabularx}
\usepackage{float}
\usepackage{epsfig}
\usepackage{epstopdf}
\usepackage{bbm}
\usepackage{mathtools,xparse}

\usepackage{epsfig} % for postscript graphics files
\usepackage{amsmath} % assumes amsmath package installed
\usepackage{amssymb}  % assumes amsmath package installed
\usepackage{bm}
\usepackage{color}
\usepackage{etoolbox}
\DeclareMathOperator*{\argmax}{arg\,max}
\DeclareMathOperator*{\argmin}{arg\,min}
\usepackage{comment}
\usepackage{diagbox}
\usepackage[ruled, lined, longend, linesnumbered]{algorithm2e}
\usepackage{booktabs,multirow}
\usepackage{bigstrut}
\usepackage{cite}

\usepackage{lipsum}
\usepackage{hyperref}
\usepackage{balance}

\usepackage{amsfonts}
\usepackage{mathtools}
\usepackage{graphicx}
\usepackage[capitalise]{cleveref}
\usepackage{hyperref}
\usepackage[nodisplayskipstretch]{setspace}
\begin{document}
\maketitle

% As a general rule, do not put math, special symbols or citations
% in the abstract or keywords.
\begin{abstract}
Safety assurance is a critical yet challenging aspect when developing self-driving technologies. Hamilton-Jacobi backward-reachability analysis is a formal verification tool for verifying the safety of dynamic systems in the presence of disturbances. However, the standard approach is too conservative to be applied to self-driving applications due to its worst-case assumption on humans' behaviors (i.e., guard against worst-case outcomes). In this work, we integrate a learning-based prediction algorithm and a game-theoretic human behavioral model to online update the conservativeness of backward-reachability analysis. We evaluate our approach using real driving data. The results show that, with reasonable assumptions on human behaviors, our approach can effectively reduce the conservativeness of the standard approach without sacrificing its safety verification ability. 
\end{abstract}

% Note that keywords are not normally used for peerreview papers.
%\begin{IEEEkeywords}

%\end{IEEEkeywords}

% For peer review papers, you can put extra information on the cover
% page as needed:
% \ifCLASSOPTIONpeerreview
% \begin{center} \bfseries EDICS Category: 3-BBND \end{center}
% \fi
%
% For peerreview papers, this IEEEtran command inserts a page break and
% creates the second title. It wipassivekyll be ignored for other modes.
\IEEEpeerreviewmaketitle

\section{Introduction}

Our society is rapidly advancing towards self-driving cars that interact with human-driven cars on public roads. Safety assurance is a critical yet challenging aspect when designing such autonomous systems. Hamilton-Jacobi (HJ) reachability analysis is a formal safety verification tool for verifying the safety of dynamic systems \cite{chen2018hamilton,bansal2017hamilton}. Under HJ-reachability paradigm, both forward \cite{althoff2014online, liu2017provably, bansal2020hamilton} and backward-reachability analysis \cite{leung2020infusing, Wang2020on, li2020prediction} can be used to ensure safety in human-robot interaction applications. For instance, forward-reachability analysis can be used to compute the human's forward-reachable set (FRS) that should be avoided by the robot. Often, the FRS leads to overly conservative robot behaviors due to its open-loop spirit. Different from forward-reachability analysis, backward-reachability analysis formulates the human-robot interaction as a differential game and computes the full backward-reachable tube (BRT) that contains states that may result in the human-robot joint system being unsafe, assuming the human is adversarial and has full control bound\footnote{An agent has full control bound if it can maneuver freely and is only constrained by the maximal bound of controls that its actuator can provide.}, within some time horizon. When operating online, the BRT can either be used as a safety monitor that switches the robot's controller to the optimal safety controller when the human is about to breach the BRT \cite{li2020prediction} or infused into the robot's low-level controller design as constraints \cite{leung2020infusing, Wang2020on}. Backward-reachability analysis is less conservative since it allows the robot to react to the human's adversarial actions. However, it is still impractical for self-driving applications due to its worst-case assumption on humans' behaviors (adversarial human with full control bound). Therefore, mitigation strategies are needed for generating a practical yet effective BRT.

Since the conservativeness of the full BRT is induced by assuming that the human-driven car utilizes its full control bound to drive the joint system to unsafe states, a natural solution is to directly limit the human-driven car's control bound. Such a technique has been exploited in both forward and backward-reachability analysis. For instance, \cite{bansal2020hamilton} assumes that the human operates under a probabilistic model parameterized by some unknown parameters and casts the human state prediction problem as a forward-reachability problem in the joint state space of the human and the belief over the unknown model parameters. When computing the human's FRS, \cite{bansal2020hamilton} limits the human's control bound based on his/her likely actions. Despite being robust to mis-specified human models and priors, \cite{bansal2020hamilton} separates the human from the robot and neglects the effects from interactions. In traffic settings, self-driving cars might get stuck with such an approach. The approach in \cite{li2020prediction} exploits a learning-based predictor to forecast the human-driven car's future trajectory and classifies the prediction into one of the pre-defined driving modes (e.g., turn left). Each driving mode is associated with a less-conservative human-driven car's control bound, and the corresponding BRT can be computed offline and queried online for safety verification. Such an approach works well if the learning-based predictor offers the human driver's future trajectories based on the self-driving car's future plan candidates, instead of only on map information and historical observations (environmental states). Without that, the estimated human-driven car's control bound can still be conservative in interactive scenarios since it doesn't account for the negotiation between the human-driven car and the self-driving car. %Learning-based predictors work well in terms of predicting the human-driven car's driving mode when only considering effects from environmental states (map information and historical observations), but they don't account sufficiently for the negotiation between the human-driven car and the self-driving car. Thus a predicted driving mode's BRT may still be conservative in interactive scenarios.

Our key insight is that: \textit{in interactive scenarios, the distribution of human-driven car's actions is not only constrained by environmental states but also affected by the human driver's internal state that governs his/her behaviors in negotiations. Estimating the human driven-car's control bound online with both environmental states and the human driver's internal state accounted for can help to obtain a less-conservative BRT.}

Overall, we make the following contributions towards practical safety verification for self-driving cars:

\vspace{+0.2cm}
\noindent
\textbf{Developing a framework for constructing a practical yet effective BRT by leveraging learning-based prediction and game-theoretic reasoning.} We propose to hierarchically estimate the human-driven car's control bound. We first utilize a learning-based method to estimate a control bound with the environmental information accounted for, and further modify the bound by reasoning about the human's willingness to cooperate in a game-theoretic fashion. Then the updated control bound is used to query an appropriate BRT online for safety verification.

\vspace{+0.2cm}
\noindent
\textbf{Verifying the effectiveness of our approach with real driving data.} We compare our approach with the standard full BRT using real traffic data. We show that our approach is significantly less conservative while maintaining safety verification ability.

\section{Preliminaries}

\subsection{Hamilton-Jacobi reachability analysis} 

%Hamilton-Jacobi (HJ) reachability analysis is a formal verification tool that verifies the safety performance of a dynamic system in the presence of disturbance by monitoring the system's reachable states. In this work, we exploit HJ backward-reachability analysis to compute the full backward-reachable tube (BRT) that contains states that may result in the dynamic system being unsafe, assuming the system is subject to worst-case disturbances, within some time horizon.
In this section, we briefly introduce the HJ backward-reachability analysis. %that is used as the safety verification tool in this work.
More details can be found in \cite{chen2018hamilton,bansal2017hamilton}.

\noindent
\textbf{Backward-reachable tube.} We consider the evolution of a dynamical system governed by the differential equation: $\dot{s}(t) = f\big(s(t), u(t), d(t)\big)$, where $s \in \mathbb{R}^{n_s}$ denotes the state of the system, $u \in \mathbb{R}^{n_u}$ denotes the bounded control, and $d \in \mathbb{R}^{n_d}$ denotes the bounded disturbance. We assume that we have access to a target set $\mathbb{G}$ that contains the unsafe states of the dynamical system. Then we define the backward-reachable tube (BRT) of the dynamical system ($\mathcal{O}_{\tau}$ ) with respect of $\mathbb{G}$ as a set of initial states from which the disturbance can drive the dynamical system to $\mathbb{G}$ within a time horizon $|\tau|$ despite the optimal control efforts, namely,
\begin{align}
    \mathcal{O}_{\tau} & = \{s_0 \in \mathbb{R}^{n_s}: \exists d(\cdot), \forall u(\cdot), \nonumber \\
    & \exists t \in [\tau, 0], s(t | f, s_0, d(\cdot), u(\cdot)) \in \mathbb{G} \},
\end{align}
where $\tau$ is a negative number since the system dynamics are propagated backwards in time.

\noindent
\textbf{Level-set method.}  The BRT $\mathcal{O}_{\tau}$ can be characterised as a zero sub-level set of a value function $V(s, \tau)$:
\begin{align}
    \mathcal{O}_{\tau} = \{s \in \mathbb{R}^{n_s}: V(s, \tau) < 0  \},
\end{align}
where the value function $V(s, t)$ is the viscosity solution of the following Hamilton-Jacobi Isaacs partial differential equation \cite{evans1984differential}:
\begin{align}
\min\{ \frac{\partial V(s, t)}{\partial t} + \min_{d \in \mathcal{D}} \max_{u \in \mathcal{U}} \nabla V(s, t)^\top f(s, u, d), \nonumber\\
V(s, 0) - V(s, t)\} = 0, t \in [\tau, 0], \label{equ: HJI PDE}
\end{align}
and $V(s, t)$ can be computed via dynamic programming \cite{mitchell2005time} with terminal condition $V(s, 0) = l(s)$, where $l(s)$ is defined as $l(s) \leq 0 \iff s \in \mathbb{G}$.

\subsection{Backward-reachability analysis in self-driving settings}

\noindent
\textbf{System dynamics.} We consider pairwise interactions between the self-driving car and each human-driven car in the environment (i.e., we verify the self-driving car's safety with respect to each human-driven car separately). Following \cite{li2020prediction}, we consider a dynamical system that encodes the relative dynamics between the self-driving car and the human-driven car in a pairwise interaction. Specifically, the dynamics of the human-driven car is represented using an extended uni-cycle model and the dynamics of the self-driving car is modeled using a high-fidelity bicycle model. The state of the relative dynamic system is defined as $s = \big(x_{rel}, y_{rel}, \psi_{rel}, v_r, v_h\big)$, where $x_{rel}$ ($y_{rel}$) is the x(y)-coordinate of the human-driven car in a coordinate frame centered at the geometric center of the self-driving car with $x$-axis aligned with the heading of the self-driving car, $\psi_{rel}$ denotes the relative heading between the two cars, and $v_r$ ($v_h$) denotes the speed of the self-driving car (human-driven car). We let $u_r(t) = (a_r(t), \delta_f(t))$ denote the control input of the self-driving car with $a_r$ denoting the acceleration and $\delta_f$ denoting the front wheel rotation, and let $u_h(t) = (a_h(t), \omega_h(t))$ denote the control input of the human-driven car with $a_h$ denoting the acceleration and $\omega_h$ denoting angular speed. The evolution of the relative system is then governed by the following differential equation $\dot{s}(t) = f\big(s(t), u_r(t), u_h(t)\big)$:
\begin{align}
   & \dot{x}_{rel} = \frac{v_r y_{rel}}{l_r}\sin(\beta_r) + v_h\cos(\psi_{rel}) - v_r\cos(\beta_r),\nonumber\\
   & \dot{y}_{rel} = -\frac{v_r x_{rel}}{l_r}\sin(\beta_r) + v_h\sin(\psi_{rel}) - v_r\sin(\beta_r),\nonumber\\
   & \dot{\psi}_{rel} = \omega_h - \frac{v_r}{l_r}\sin(\beta_r ),\nonumber\\
   & \dot{v}_r = a_r,\nonumber\\
   & \dot{v}_h = a_h,
   \label{equ: relative dynamics}
\end{align}
where $l_f$ ($l_r$) denotes the front (rear) axle length of the self-driving car and $\beta_r$ is computed via $\beta_r = \tan^{-1}(\frac{l_r}{l_r+l_f}\tan(\delta_f))$. %\tr{you should explain that in this system, the human driver's action will be treated as disturbance. Otherwise, the "disturbance bound" later might be confusing for readers who don't know the details of BRT.}

\noindent
\textbf{BRT as a safety monitor.} We treat the human-driven car's control as disturbance of the relative dynamical system \eqref{equ: relative dynamics}, then we can compute the BRT that contains initial relative states that could potentially lead to collisions. Such a BRT can be used as a safety monitor that switches the self-driving car's low-level controller to the optimal safety controller when the current relative state is within the BRT \cite{li2020prediction} or can be infused into the self-driving car's low-level controller design as constraints \cite{leung2020infusing, Wang2020on}.

\noindent
\textbf{Disturbance bound.} It can be observed from \eqref{equ: relative dynamics} and \eqref{equ: HJI PDE} that the conservativeness of the BRT highly depends on the human-driven car's control bound $\mathcal{D}$. The BRT computed with the full human-driven car's control bound is overly conservative in general.
% \tr{I deleted the remaining part since it is duplicating.}%In previous work \cite{leung2020infusing, Wang2020on}, the BRT is computed using a bound that reflects the human-driven car's actuator limitation (worst-case disturbances) and is exploited without any updates during operation. We argue that the human-driven car's control bound is constrained by the road environment and the driver's own internal state. By updating the the human-driven car's control bound online via observations, we can enable the self-driving car to react to the learned knowledge and update its safety monitor to reduce conservativeness.

%\vspace{-0.3cm}
\section{Negotiation-Aware BRT} 
In this section, we propose a negotiation-aware BRT computation approach that estimates the human-driven car's control bound based not only on environmental information but also on the potential negotiation with the self-driving car. We propose to estimate such a bound hierarchically: 1) first utilize a learning-based method to estimate a control bound with the environmental information accounted for, and further modify the bound by reasoning about the internal state of the human driver in a game-theoretic fashion.
%illustrate our approach to constructing a less-conservative while safety-preserving BRT for verifying safety of self-driving cars in highly interactive scenarios where negotiations between agents are needed.

%\tr{Rename Section II as Preliminaries, and put current Section II as Section II-A, then put current Section III-A as Section II-B.}

\subsection{Estimate disturbance bound via learning-based predictor}\label{sec: control bound estimation via prediction}

As in \cite{li2020prediction}, one way to obtain a practical estimate on $\mathcal{D}$ is to leverage a learning-based prediction algorithm. Typically, a learning-based prediction algorithm will generate the human-driven car's most-likely future trajectory and the statistical prediction error given the map information and the joint history states \cite{mozaffari2020deep}. Hence, in the first layer, we utilize the most-likely trajectory and the prediction error to generate an initial control bound estimation. More specifically, we construct the Frenet frame taking the most-likely path as the reference path. In the Frenet frame, we can obtain an acceleration bound $\mathcal{D}_{a} = [a_{lower},a_{upper}]$ from the most-likely trajectory and an angular speed bound $\mathcal{D}_{\omega} = [0,0]$. The statistical prediction error\footnote{Such statistical results can be obtained on the validation sets of the learning algorithms.} can further provide additional safety margins to improve the predicted control bound's robustness in the presence of long-tailing problems in learning-based approaches.

%for $\mathcal{D}_{a}$ and $\mathcal{D}_{\omega}$.

%Note that our method distinguished from \cite{li2020prediction} by directly estimating the control bounds in the Frenet Frame with the prediction error accounted for, which improves its robustness in the presence of long-tailing problems in learning-based approaches. 

%since \cite{li2020prediction} classifies the predicted trajectory into a pre-defined driving mode to get the corresponding human-driven car's control bound, while ours directly estimate the control bounds by considering the prediction error, which improves its robustness in the presence of long-tailing problems in learning-based approaches. 

\subsection{Negotiation modeling via game-theoretic reasoning}\label{sec: negotiation modeling}

One limitation of the first-layer's estimation approach is that the obtained control bound does not account for negotiations between agents (e.g., a human will slow down if a self-driving car gradually merges in front of him/her \cite{sadigh2016planning}). Such a limitation can also lead to a conservative BRT when active maneuvers are needed in interactive scenarios. To tackle this problem, in the second layer, we exploit a game-theoretic approach to explicitly reason about negotiations and correct the predicted human-driven car's acceleration bound. We emphasize that in the second layer, the game-theoretic approach is exploited for its ability to model intense interactions between agents, mimic humans' decision-making behaviors, and reveal their internal states.

\noindent
\textbf{Assumptions and simplifications.}
We assume that the human reacts to the self-driving car via acceleration in the Frenet frame described in \cref{sec: control bound estimation via prediction}. In other words, when modeling negotiations, we assume that the human-drive car will operate along its predicted most-likely path (reference path in the Frenet frame). Such an assumption is reasonable in light of two observations: 1) learning-based predictors predict a decent potential path of a human-driven car in structured roads, as they explicitly encode features of road environments during training; 2) a human driver tends to follow a reference path and react to other drivers mainly by adjusting his/her acceleration. Since the current motion plan of the self-driving car is available in the context of safety verification, we further assume that the self-driving car operates along its current planned path when modeling negotiations (such an assumption limits the self-driving car's mobility and tends to make the human's acceleration bound prediction more conservative, but it provides a computational gain by making the negotiation modeling online usable).

\noindent
\textbf{Negotiation as a Stackelberg game.} With the above assumptions, we model the negotiation between the human-driven car and the self-driving car as a Stackelberg game \cite{bacsar1998dynamic} since it explicitly considers one player's advantages over the other player, which can be used to model different roles in a traffic negotiation. The dynamics of the game is governed by $s_{t+1} = g(s_{t}, a^{r}_t,  a^{h}_t)$, where the subscript denotes a discrete time step, $a$ denotes acceleration, and the dynamics function $g$ governs the state evolution along agents' respective paths. We assume that both the human-driven car and the self-driving car have a finite set of acceleration controllers in their negotiation. A controller $\pi \in \Pi$ defines an agent's controls for a short planning horizon $T$. At a time step $t$, an agent's controller $\pi$ maps a time increment $\tau$  ($\tau\in [0,T]$) to the acceleration command at $t\Delta_t + \tau$, where $\Delta t$ is the time integral between successive time steps,

In a negotiation, the human-driven car can either be a \textit{follower} who is willing to cooperate or be a \textit{leader} who aims to dominate the interaction. Specifically, a follower maximizes his/her own reward function while accommodating the self-driving car. Assuming that humans are noisy-rational, then a follower selects controller in response to the self-driving car's current motion plan according to the following quantal response model:
\begin{align}
   & \mathbb{P}(\pi_{h} | s_t, \pi_r^g) \propto \exp(\beta Q_t^{h}(s_t, \pi_h, \pi_r^g)),  \label{equ: follower policy}\\
   & Q_t^{h}(s_t, \pi_h, \pi_r^g) = \sum_{n = 0}^{T/ \Delta t }\hspace{-0.2cm} R_{h}(s_{t+n}, \pi_h(n\Delta t), \pi_r^g(n\Delta t)),\label{equ: follower Q}
\end{align}
%\vspace{-0.4cm}

\noindent
where $\pi^r_g$ denotes the self-driving car's current planned controls along its desired path, and $R_h$ is human's reward function learned via Inverse Reinforcement Learning \cite{ziebart2008maximum, sadigh2016planning}. In contrast to a follower, a leader makes decisions assuming the self-driving car is a follower who optimally responds to his/her controller:
\begin{align}
    \hspace{-0.2cm} \mathbb{P}(\pi_{h} | s_t) \propto \exp(\beta Q^t_{h}(s_t, \pi_h, \argmax_{\pi_r \in \Pi} Q^t_{r}(s_t, \pi_h, \pi_r)), \label{equ: leader policy}
\end{align}
%\vspace{-0.4cm}

\noindent
where the expression of $Q^t_r$ is defined analogously to \eqref{equ: follower Q} with reward function $R_r$.

%It can be observed from \eqref{equ: follower policy} and \eqref{equ: leader policy} that a follower human-driven car accommodates the self-driving car and is likely to be negotiated, while a leader human-driven car dominates the interaction and is unwillingly to be negotiated. In previous works \cite{sadigh2016information, sadigh2016planning}, the human-driven car is assumed to be always a follower when predicting the human's future states, though assigning the leading-following role a \textit{priori} may result in safety critical issues. Different from \cite{sadigh2016information, sadigh2016planning}, 

\noindent
\textbf{Leading-following role inference.} Since the leading-following role of the human-driven car in a negotiation is unknown, we treat it as an internal state that needs to be inferred online. Normally the self-driving car's sensing module runs at a much higher frequency than the safety verification module. We let $\xi_t$ denote the collection of controls of the human-driven car in the previous $T[s]$ observed by the self-driving car. We let $\theta \in \Theta = \{l,f\}$ denote the human-driven car's leading-following role and define $b_t(\theta) = \mathbb{P}(\theta^* = \theta)$ as the belief distribution over $\Theta$ at step $t$. Then the self-driving car can update the human-driven car's role via Bayesian inference:
\begin{align}\label{equ: bayesain}
    b_{t}(\theta) \propto \mathbb{Z}(\xi_t | \theta)  b_{t-1}(\theta),
\end{align}
%\vspace{-0.4cm}

\noindent
where $\mathbb{Z}(\xi_t | \theta)$ is an observation model that specifies the probability of observing $\xi_t$ if the human-driven car's role were $\theta$. We define a distance measure $d$ that evaluates the accumulative difference (Euclidean distance) between the observed human-driven car's controls ($\xi_t$) and controls obtained from a controller $\pi \in \Pi$ evaluated at the times steps that the observations were made, and use $d$ to find the controller that best matches the observed controls. Then the observation model is defined as:

\noindent
\small
\begin{align}\label{equ: observer}
    \hspace{-0.2cm}\mathbb{Z}(\xi_t | \theta) =
    \begin{cases}
    \mathbb{P}\big( \argmin_{\pi_h \in \Pi} d(\pi_h,\xi_t) | s_{t-1}, \pi_r^g\big), \text{if } \theta = f,\\
    \mathbb{P}\big( \argmin_{\pi_h \in \Pi} d(\pi_h,\xi_t) | s_{t-1}\big), \text{if } \theta = l,
    \end{cases}
\end{align}
\normalsize
%\vspace{-0.3cm}

\noindent
where the probabilities are computed using \eqref{equ: follower policy} and \eqref{equ: leader policy}. With \eqref{equ: bayesain} and \eqref{equ: observer}, we can ``correct" the human's predicted acceleration bound from the learning-based predictor by explicitly modeling and inferring the human's willingness to be negotiated.

\subsection{Online update the backward-reachable tube.}

Since $\Pi$ is finite, we have a finite set of potential acceleration bounds ($\bar{\mathcal{D}}_a$) of the human-driven car associated with the controllers. For each acceleration bound $\mathcal{D}_a^i$, we offline compute a BRT, $\mathbb{O}^{i}_{\tau}$, using the control bound $(\mathcal{D}_a^i, \mathcal{D}_\omega)$, where $\mathcal{D}_\omega$ denotes the human-driven car's angular speed bound obtained from the learning-based predictor with safety margin described in \cref{sec: control bound estimation via prediction}. When operating online, at each step, we first update $b_t(\theta)$ based on the current observation, then run the following algorithm to augment BRTs based on the distribution over the human-driven car's future control bounds with respect to $b_t(\theta)$. The notation $\Delta$ denotes a desired confidence bound (we aim to augment BRTs such that their marginal probability is larger than $\Delta$). When $\Delta = 1$, the resulting BRT is equivalent to the one that considers the whole acceleration bound (most conservative). Note that the obtained BRT is in the Frenet frame described in \cref{sec: control bound estimation via prediction}.

%\vspace{-0.4cm}
\begin{algorithm}[ht]
        \caption{Update BRT}
        \label{alg: Algorithm}
            
            For each controller $\pi_h^i \in \Pi$, compute its expected probability $\mathbb{P}(\pi_h^i)$ with respect to $b_t(\theta)$;
            
            Sort $\Pi$ with respect to $\mathbb{P}(\pi_h^i)$ in descending order;
        
            Initialize $\mathbb{O}^{*}_{\tau} \leftarrow \emptyset$, $P = 0$;
            
            \For{$i = 1,2,...,|\Pi|$}{
                  $P = P + \mathbb{P}(\pi_h^i)$;
                  
                  $\mathbb{O}^{*}_{\tau} \leftarrow \mathbb{O}^{*}_{\tau} \cup \mathbb{O}^{i}_{\tau}$;
                   
                  \If {$P >= \Delta$}{
                        break;
                  }
            }
            Return  $\mathbb{O}^{*}_{\tau}$;
\end{algorithm}
%\vspace{-0.6cm}

\section{Experiment and Result}

\subsection{Experiment design}

We use a controller-switching strategy for ensuring the self-driving car's safety. We extracted 15 pairs of two-car interaction trajectories at an uncontrolled roundabout from \cite{interactiondataset}. For each interaction log, we use a BRT to ensure the safety performance of one selected car (treated as the self-driving car) when it interacts with the other car (treated as the human-driven car). Specifically, both cars are set to execute their original controls from the log. When the human-driven car breaches the self-driving car's BRT, the interaction is identified as unsafe, and a safety controller overrides the self-driving car's controls.

\noindent
\textbf{Manipulated variables.} The manipulated variable is the choice of different algorithms for generating a BRT for safety verification. In addition to our approach that online updates the BRT using the proposed hierarchical method, we also use the full BRT (the one that assumes the worst outcomes) in standard backward-reachability analysis and the prediction-based BRT \cite{li2020prediction} as baseline algorithms.

\noindent
\textbf{Dependent measures.} Since all the extracted interactions are safe, we quantify the conservativeness and effectiveness of a BRT using two measures: relative speed and relative distance between two cars when the human-driven car breaches the self-driving car's BRT.

\noindent
\textbf{Hypothesis.} Estimating the human driven-car's control bound online with both environmental states and the human driver's internal state accounted for can effectively reduce the conservativeness of the BRT without sacrificing its safety verification ability.

\subsection{Implementation details.}
\begin{figure}[ht]
\begin{center}
\begin{picture}(300, 120)
%%%%%%%%%%%%%%%%%%%%%%%%%%%%%%
\put(0,  0){\epsfig{file=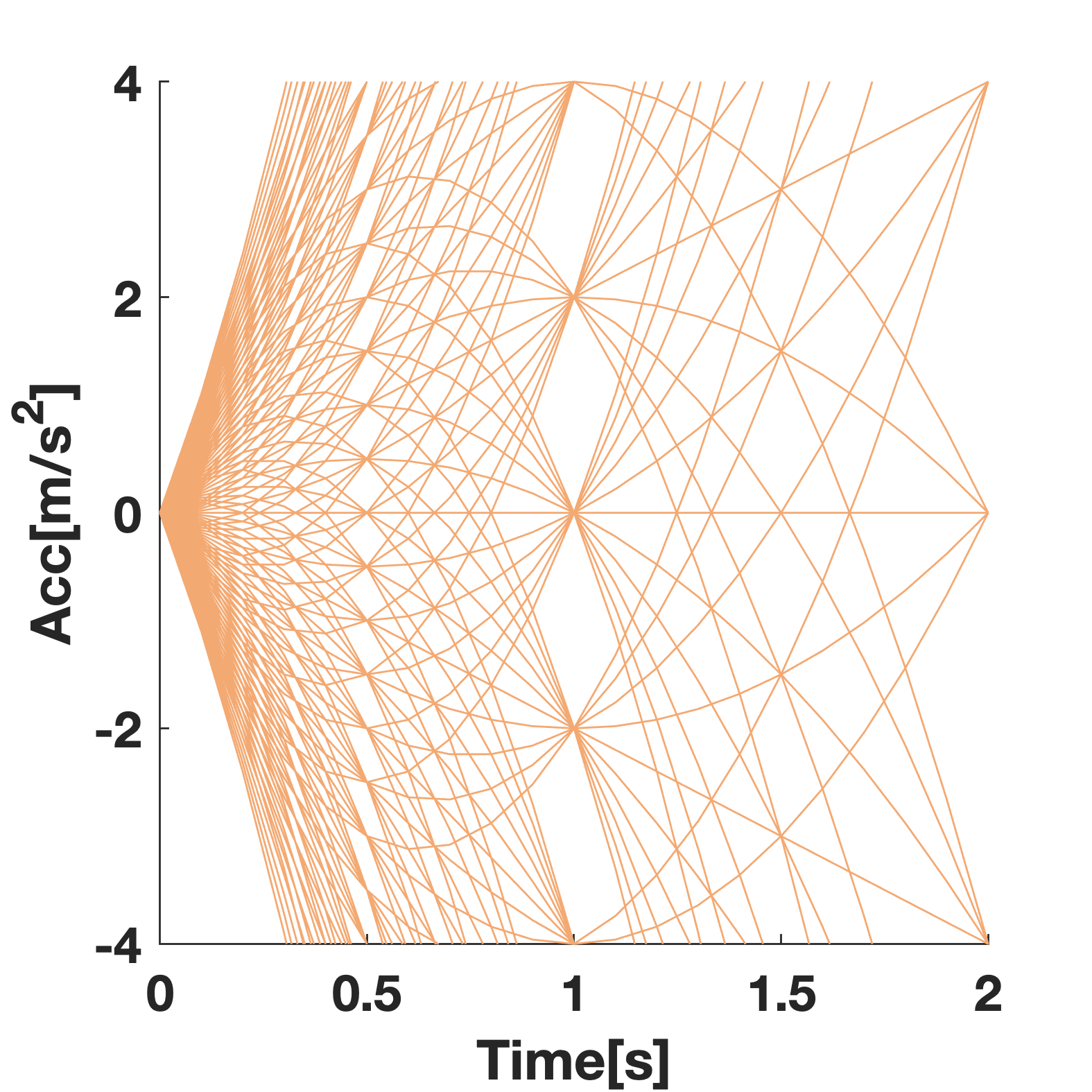, width = 0.47\linewidth, trim=0.0cm 0.0cm 0.4cm 0.5cm,clip}}  %%%
\put(120,  0){\epsfig{file=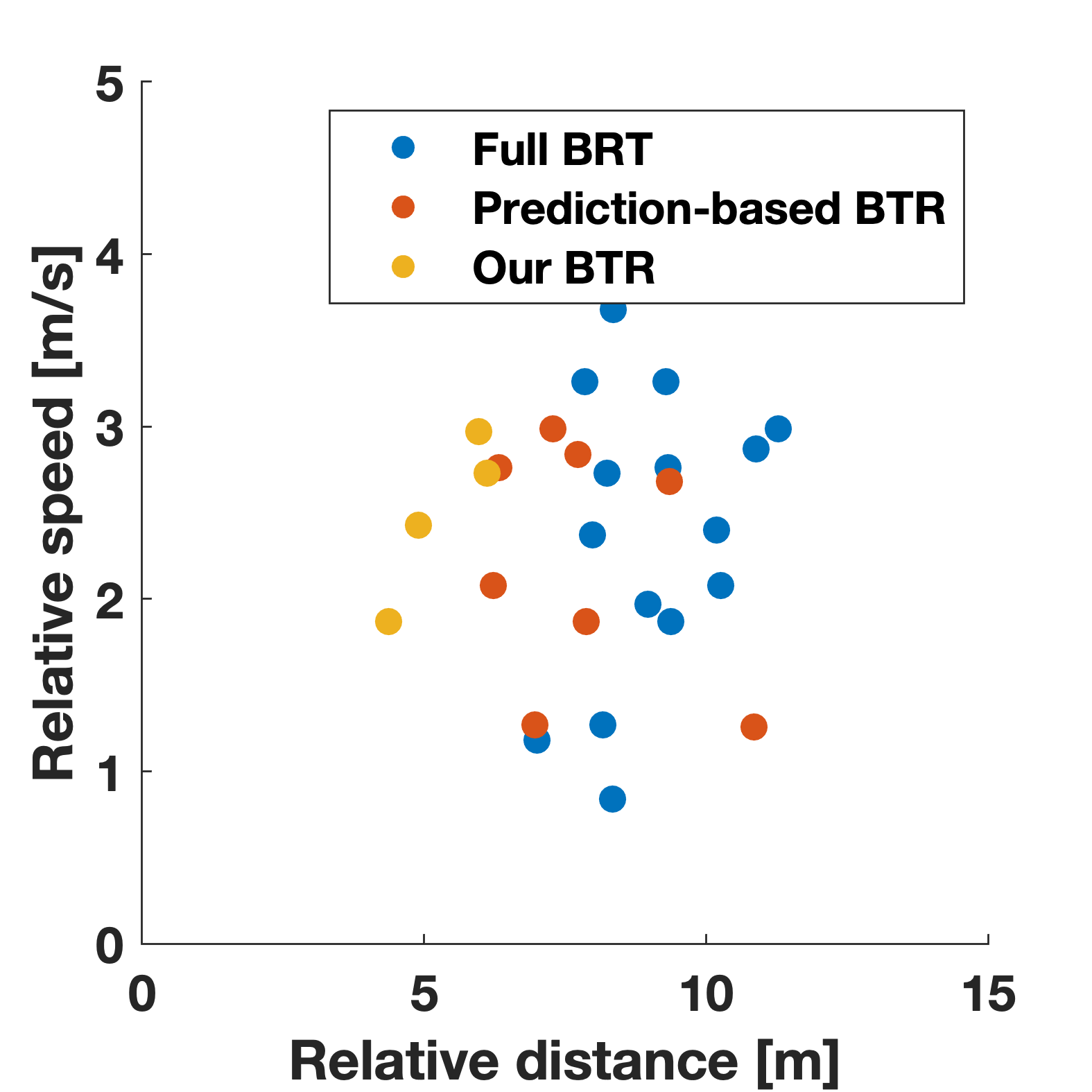, width = 0.47\linewidth, trim=0.0cm 0.0cm 0.4cm 0.5cm,clip}}  %%%
\end{picture}
\end{center}
%\vspace{-0.5cm}
\caption{Left: sampled acceleration controllers. Right: evaluation result.}
\label{fig: eval_result}
\vspace{-0.4cm}
\end{figure}

We model the acceleration controller in the negotiation modeling (\cref{sec: negotiation modeling}) as a second-order polynomial and sample 200 acceleration controllers to construct $\Pi$ (\cref{fig: eval_result}). When evaluating \eqref{equ: follower policy} and \eqref{equ: leader policy}, we use multi-thread computing to evaluate each controller in parallel so that the negotiation modeling is real-time feasible. The planning horizon $T$ is $2s$, and the confidence bound in \cref{alg: Algorithm} is $\Delta = 0.9$. We initialize the distribution over the human-driven car's leading-following role as a uniform distribution over $\Theta$. The learning-based prediction algorithm used in this work is adopted from \cite{chai2019multipath} and trained in \cite{interactiondataset}.

%\vspace{+0.2cm}

\subsection{Result}

\begin{figure}[ht]
\begin{center}
\begin{picture}(300, 350)
%%%%%%%%%%%%%%%%%%%%%%%%%%%%%%
\put(0,  220){\epsfig{file=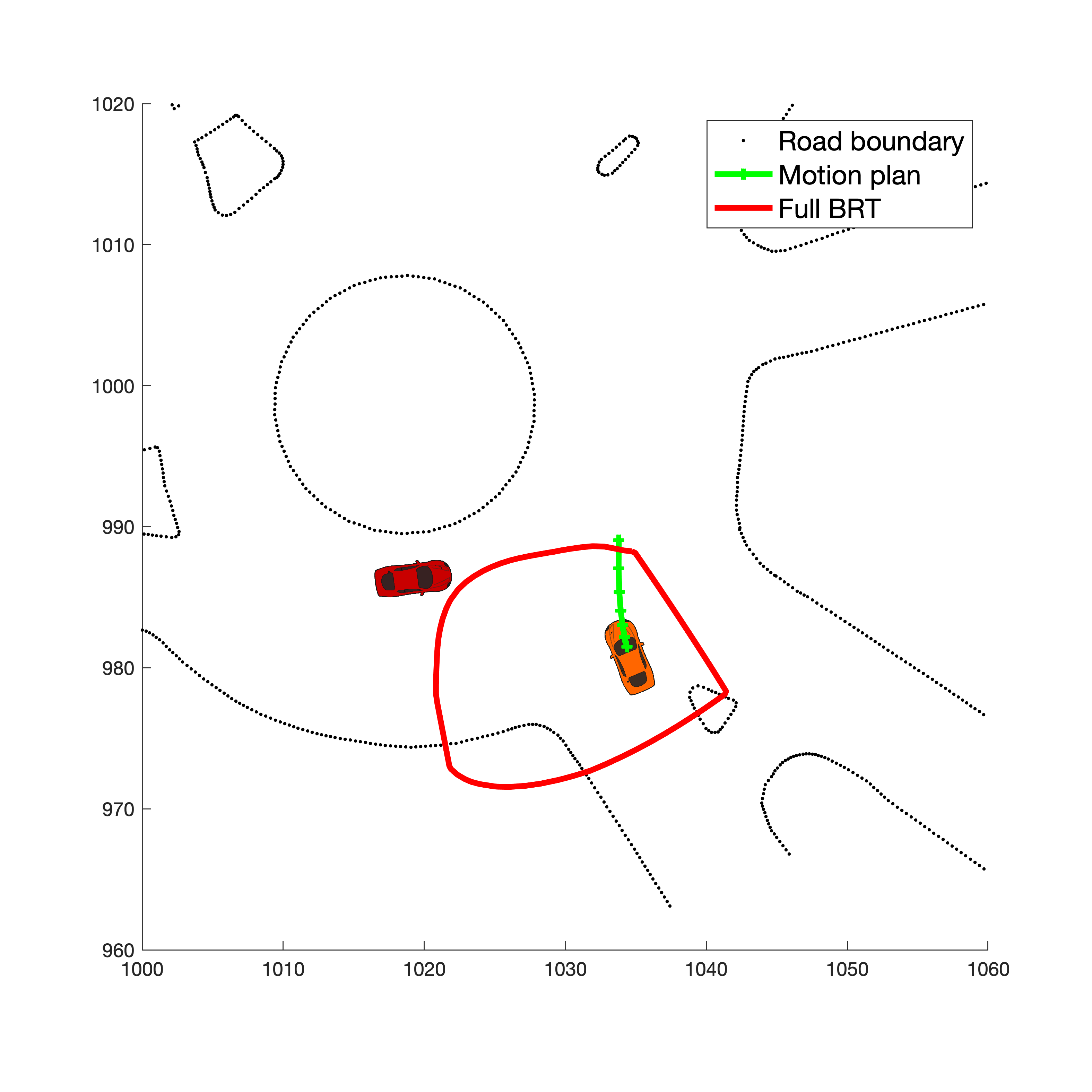, width = 0.5\linewidth, trim=2.5cm 2.5cm 0.5cm 0.5cm,clip}}  %%%
\put(120,  220){\epsfig{file=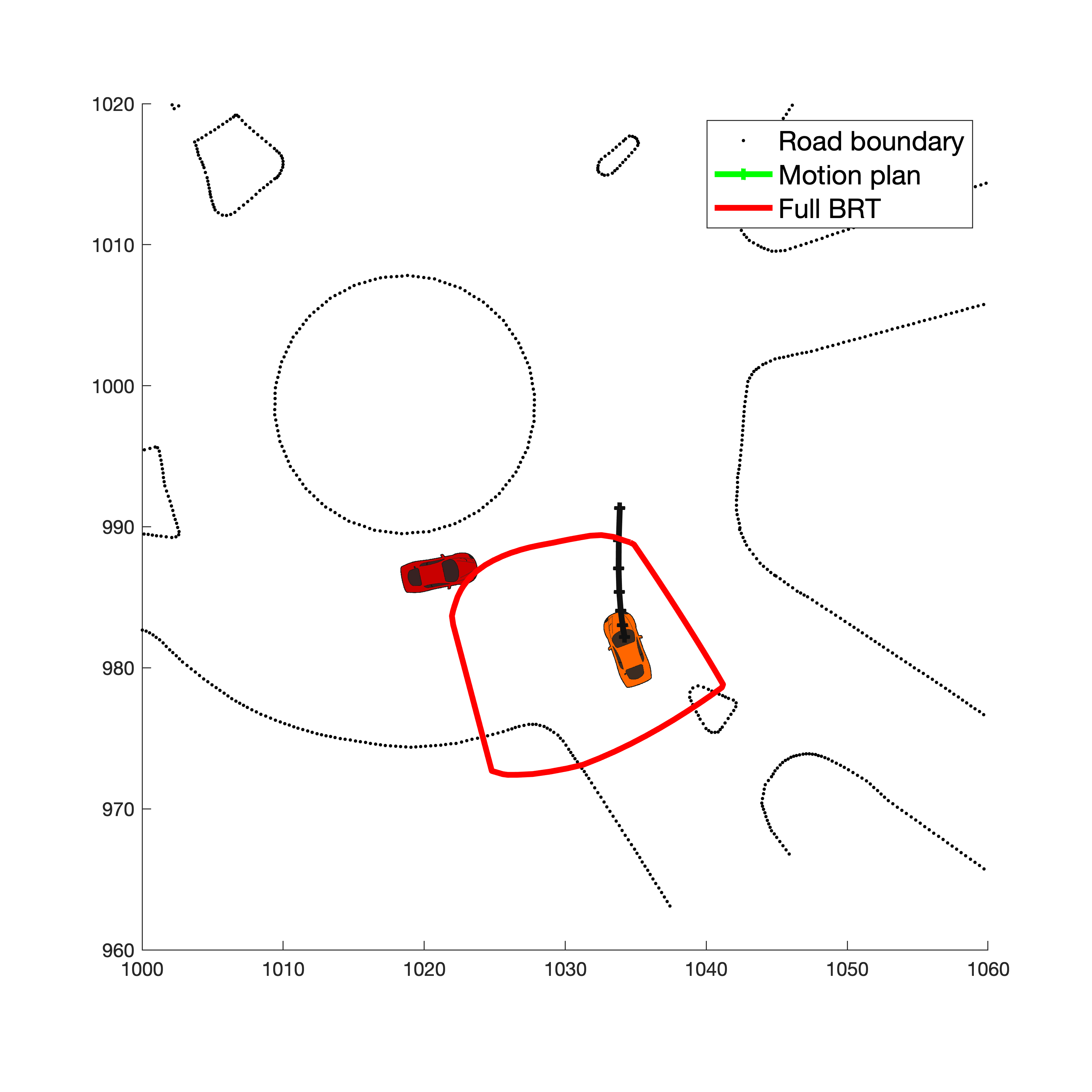, width = 0.5\linewidth, trim=2.5cm 2.5cm 0.5cm 0.5cm,clip}}  %%%

\put(0,  110){\epsfig{file=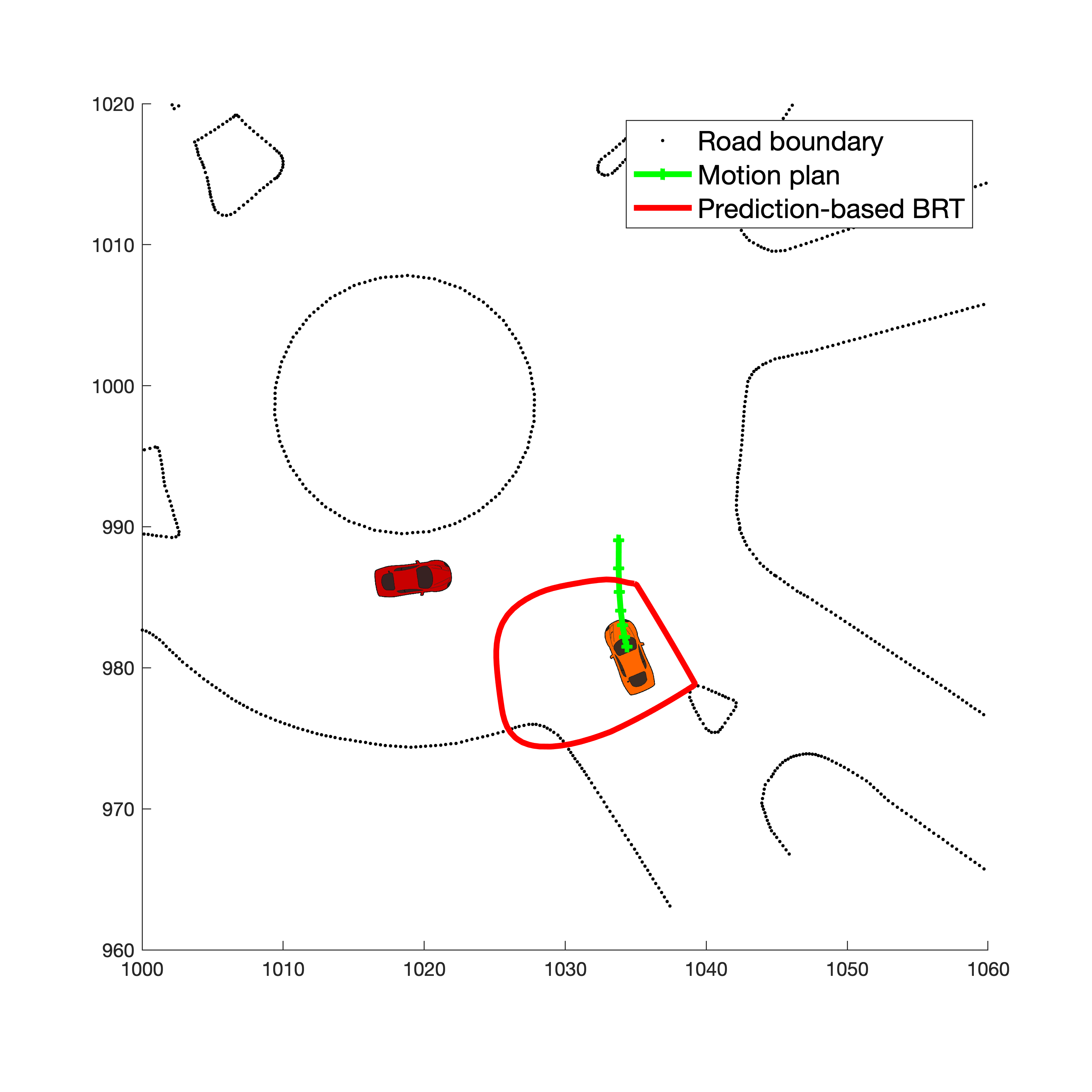, width = 0.5\linewidth, trim=2.5cm 2.5cm 0.5cm 2.5cm,clip}}  %%%
\put(120,  110){\epsfig{file=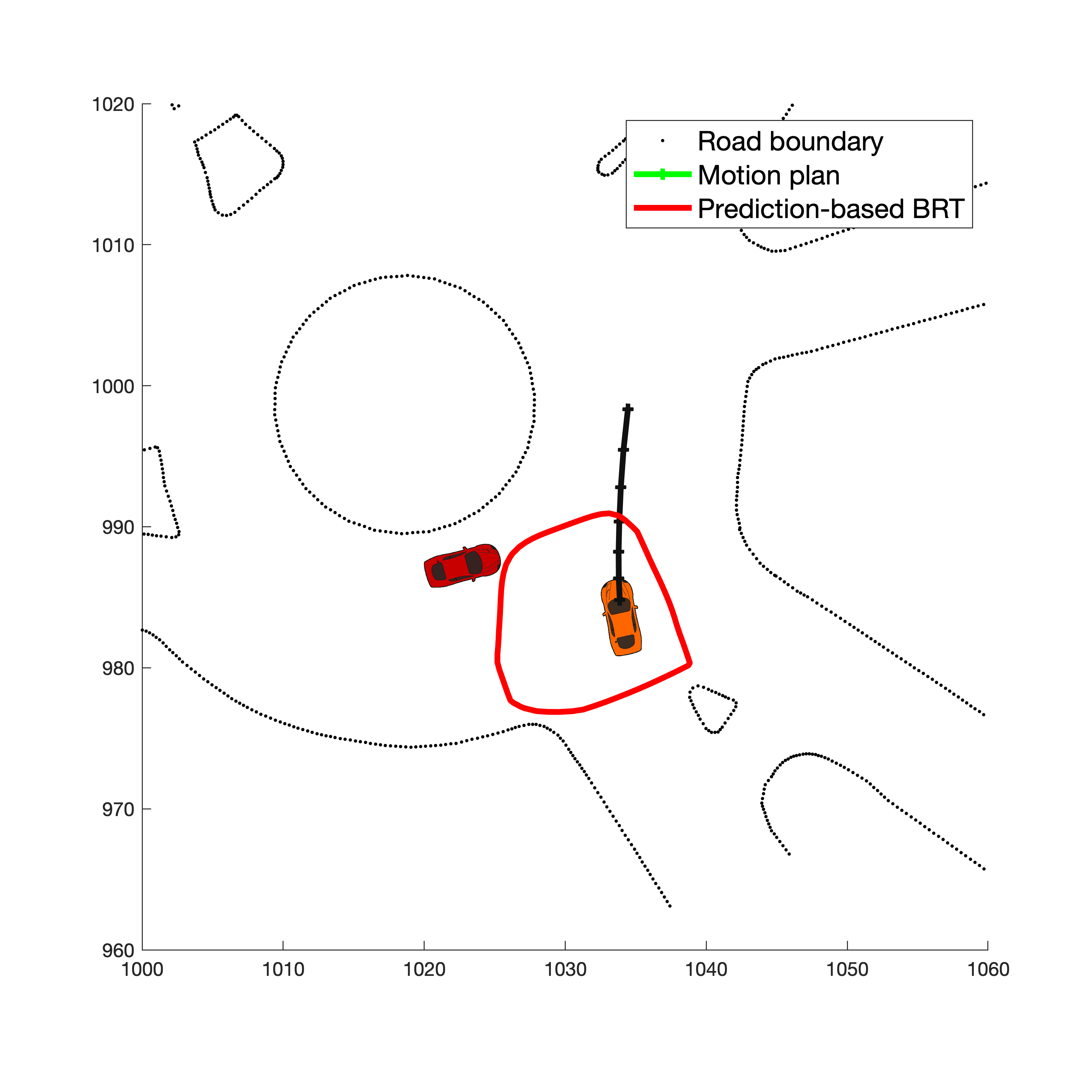, width = 0.5\linewidth, trim=2.5cm 2.5cm 0.5cm 2.5cm,clip}}  %%%

\put(0,  0){\epsfig{file=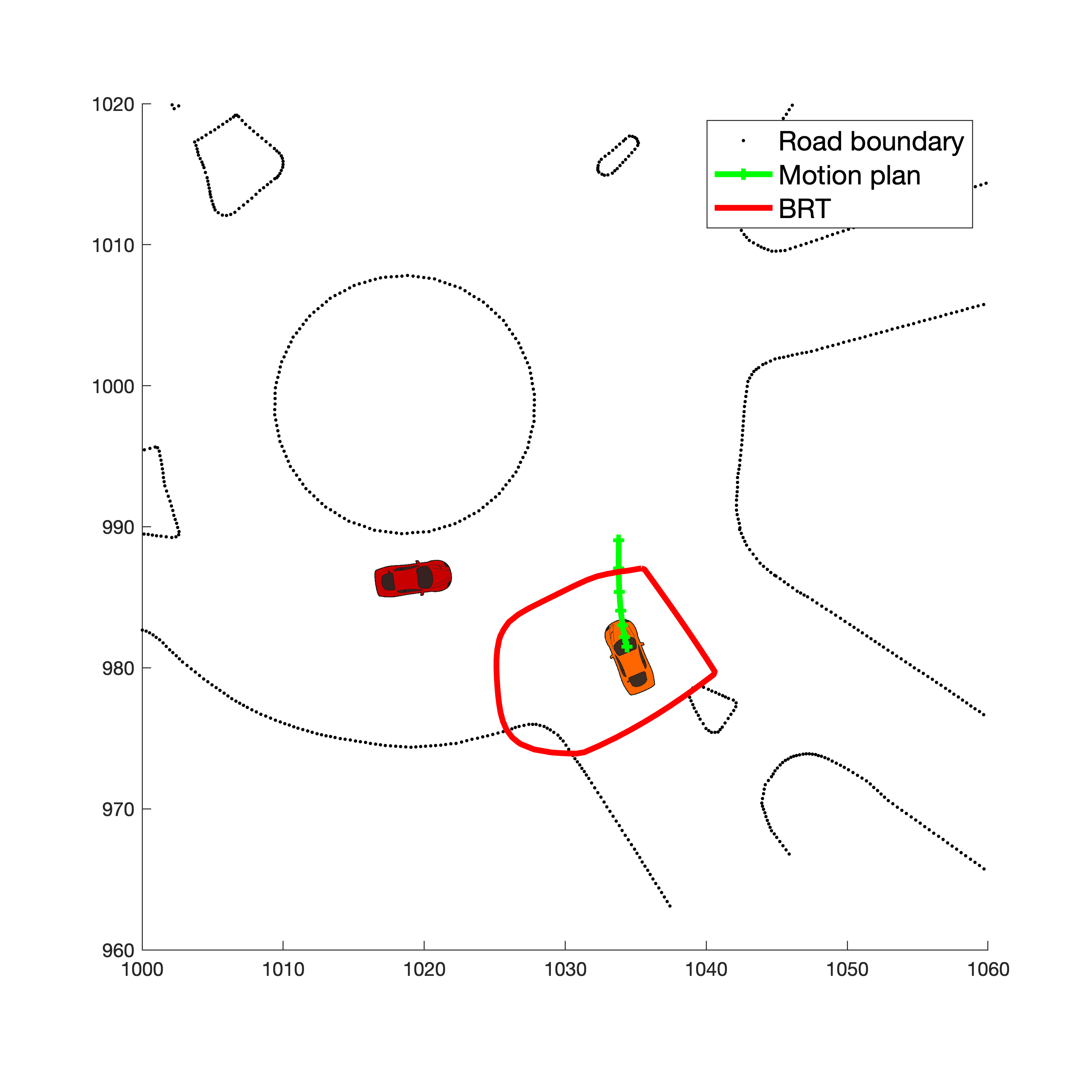, width = 0.5\linewidth, trim=2.5cm 2.5cm 0.5cm 2.5cm,clip}}  %%%
\put(120,  0){\epsfig{file=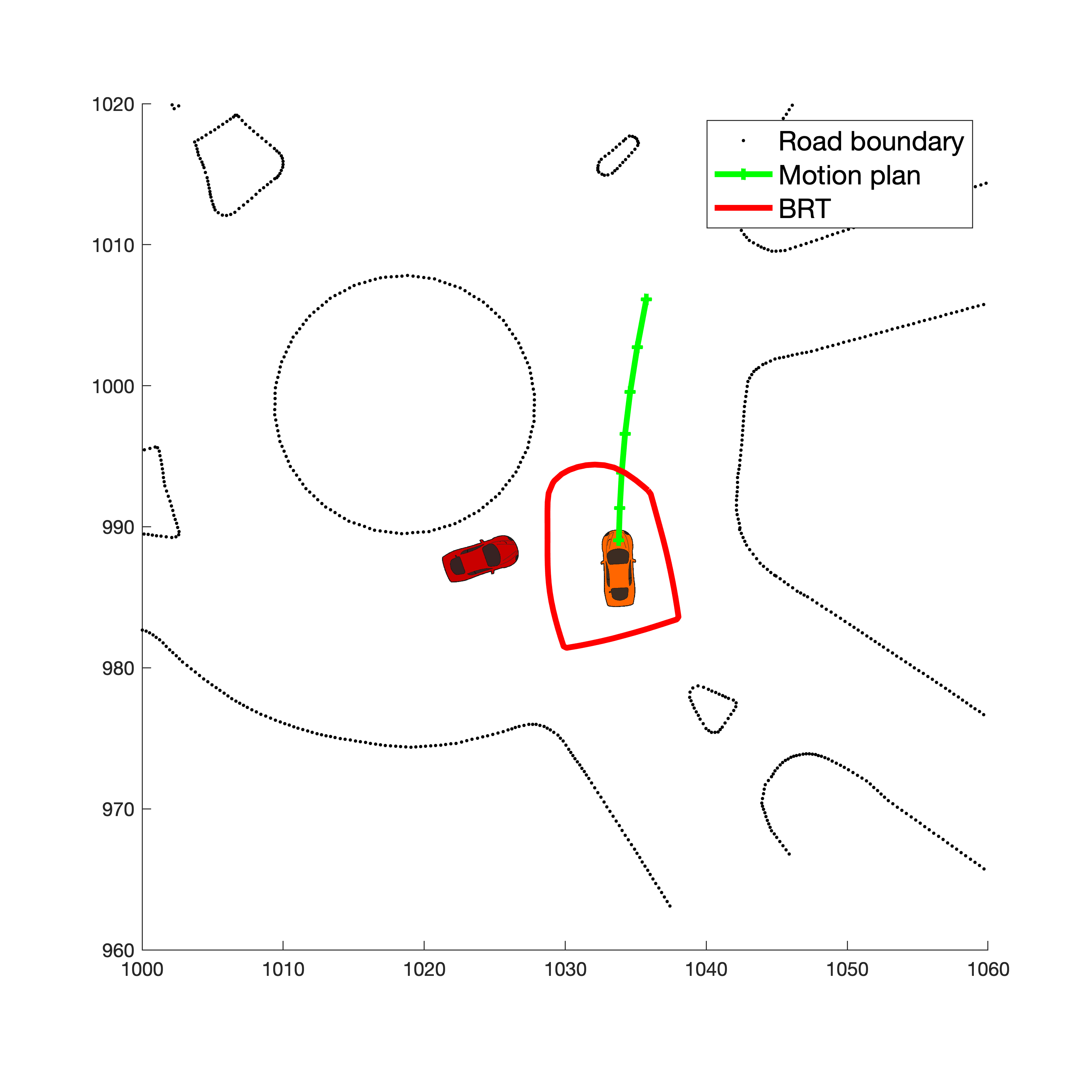, width = 0.5\linewidth, trim=2.5cm 2.5cm 0.5cm 2.5cm,clip}}  %%%
\put(0,320){(a-1)}
\put(0,210){(b-1)}
\put(0,100){(c-1)}
\put(120,320){(a-2)}
\put(120,210){(b-2)}
\put(120,100){(c-2)}

\end{picture}
\end{center}
%\vspace{-0.5cm}
\caption{(a-1)-(a-2) show two sequential steps in an experiment with the full BRT. (b-1)-(b-2) and (c-1)-(c-2) show steps in similar simulations with the BRT generated by the prediction-based reachability analysis \cite{li2020prediction} and our approach, respectively. The BRTs showed in the cartesian frame above are projected from the Frenet frame.}
\label{fig: demo}
%\vspace{-0.4cm}
\end{figure}
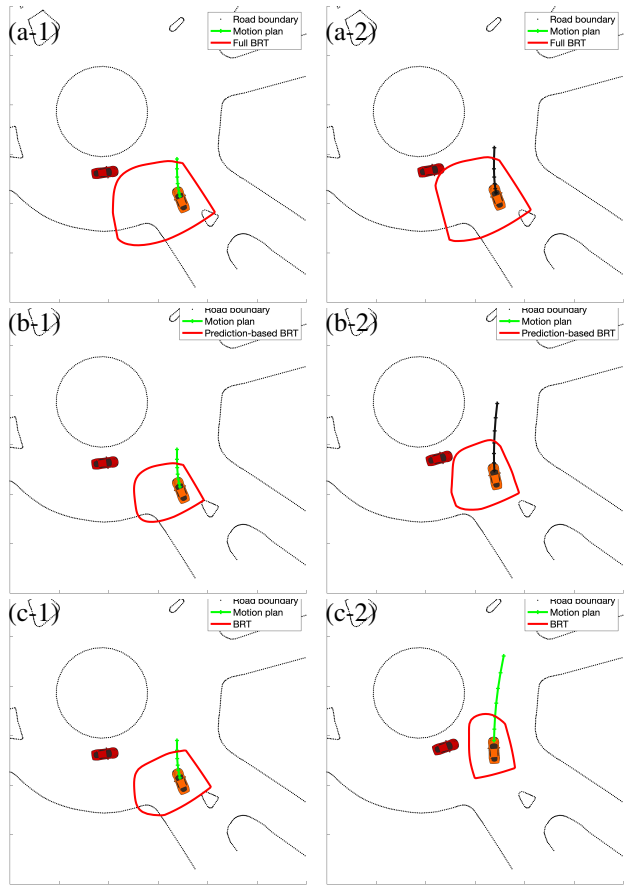

In the right plot of \cref{fig: eval_result}, we show the relative speed/distance between two cars when the human-driven car breaches the self-driving car’s BRT. It can be observed that, even though all 15 interactions are safe, the human-driven car breaches the full BRT in all the interactions in our experiment, which shows the conservativeness of the full BRT. With the prediction-based BRT, 9 interactions are identified as unsafe interactions, which indicates that leveraging end-to-end trajectory prediction can reduce some of the conservativeness of the full BRT. With our approach, only 4 interactions are identified as unsafe interactions by the BRT. Moreover, the average ground-truth minimal time-to-collision (TTC) \cite{jansson2005collision} of the identified unsafe interactions is $TTC=2.4s$ \footnote{Typically, a two-car interaction with less than 2[s] of TTC can be viewed as a dangerous interaction}, which shows that our approach to dynamically updating the BTR further reduces the conservativeness of the full BRT while maintaining the safety verification ability of backward-reachability analysis. In \cref{fig: demo}, we show an instance of our experiment in a strong negotiation. In (a-1)-(a-2) and (b-1)-(b-2), both the full BRT and the prediction-based BRT identify the interaction as unsafe, and the safety controller stops the self-driving car, although the human-driven car is trying to yield and stops at $6 [m]$ away from the self-driving car. In contrast, our approach identifies that the human-driven car is willing to cooperate and quickly adjusts the BRT to ensure a safe and effective interaction. The preliminary experiment result shows that, indeed, estimating the human driven-car's control bound online with both environmental states and the human driver's internal state accounted for can help to obtain a less-conservative while safety-preserving BRT.

\section{Conclusion}

\noindent
\textbf{Summary.} Safety assurance is a critical yet challenging aspect when developing self-driving technologies. Hamilton-Jacobi backward-reachability analysis is a formal verification tool for verifying the safety of dynamic systems in the presence of disturbances. However, the standard approach is too conservative to be applied to self-driving applications due to its worst-case assumption on humans' behaviors. In this work, we integrated a learning-based prediction algorithm and a game-theoretic negotiation model to online update the conservativeness of backward-reachability analysis. We evaluated our approach using real driving data. The results showed that, with reasonable assumptions on human behaviors, our approach can effectively reduce the conservativeness of the standard approach without sacrificing its safety verification ability.

\noindent
\textbf{Limitations and future works.} Our work is limited in many ways. One of the limitations is the simplifications made in the negotiation modeling (i.e., assuming agents operate along pre-defined paths using pre-defined acceleration controllers). Although such simplifications may affect the accuracy of the control bound estimation, they make integrating learning-based prediction and game-theoretic reasoning more practical, and they provide computational benefits that make the negotiation modeling online usable. Besides, the BRT augmentation step in \cref{alg: Algorithm} provides some robustness to the effects induced by these simplifications. An alternative approach is to solve the full Stackelberg game with discretized state-action space offline \cite{fisac2019hierarchical} and use the pre-computed leader and follower policies for online inference. We only compared our approach with the baselines in a signal traffic scenario using a small number of traffic interactions. More comprehensive experiments are needed to validate our approach in various traffic scenarios. In this work, we used a controller-switching strategy when using the BRT to ensure the safety performance of the self-driving car. Further experiments are needed to validate our approach's advantages over the baselines in a setting where the BRT is infused into the low-level controller's design process \cite{leung2020infusing}.

\bibliographystyle{IEEEtran}

%%\bibliography{ref} C:\Users\hp\Documents
%\small
\balance
\bibliography{Ref}

\end{document}